\pdfoutput=1
\documentclass[11pt]{article}

\usepackage[final]{acl}
\usepackage{times}
\usepackage{latexsym}

\usepackage{enumitem}

\usepackage[T1]{fontenc}
\usepackage[utf8]{inputenc}

\usepackage{microtype}

\usepackage{inconsolata}

\usepackage{graphicx}

\usepackage{amsmath}
\usepackage{amssymb}
\usepackage{mathtools}
\usepackage{amsthm}
\usepackage{amsfonts}
\usepackage{multirow}
\usepackage{booktabs}
\usepackage{makecell}
\usepackage[most]{tcolorbox}
\usepackage{algorithm}
\usepackage{algorithmic}

\newcommand{\ModelName}{{straQ*}}
\newcommand{\vanillaFT}{{SFT}}

\title{Convert Language Model into a Value-based Strategic Planner}
\author{
  \textbf{Xiaoyu Wang\textsuperscript{1,2}\thanks{Work was done during the internship at Geely.}},
  \textbf{Yue Zhao\textsuperscript{1}},
  \textbf{Qingqing Gu\textsuperscript{1}},
  \textbf{Zhonglin Jiang \textsuperscript{1}},\\
  \textbf{Xiaokai Chen\textsuperscript{2}},
  \textbf{Yong Chen\textsuperscript{1}},
  \textbf{Luo Ji\textsuperscript{1}\thanks{Corresponding Author.}}
\\
\\
  \textsuperscript{1} Geely AI Lab, Beijing, China \\
  \textsuperscript{2} Beijing Institute of Technology, Beijing, China
\\
  \small{
    \textbf{Correspondence:} \href{mailto:email@domain}{Luo.Ji1@geely.com}
  }
}

\begin{document}

\maketitle

\begin{abstract}
Emotional support conversation (ESC) aims to alleviate the emotional distress of individuals through effective conversations. Although large language models (LLMs) have obtained remarkable progress on ESC, most of these studies might not define the diagram from the state model perspective, therefore providing a suboptimal solution for long-term satisfaction. To address such an issue, we leverage the Q-learning on LLMs, and propose a framework called straQ*. Our framework allows a plug-and-play LLM to bootstrap the planning during ESC, determine the optimal strategy based on long-term returns, and finally guide the LLM to response. Substantial experiments on ESC datasets suggest that straQ* outperforms many baselines, including direct inference, self-refine, chain of thought, finetuning, and finite state machines. Our implementation is available at \url{https://github.com/suran662/StraQ}.
\end{abstract}

\section{Introduction}

Emotional Support Conversation (ESC) refers to dialogues aimed at alleviating a seeker's emotional distress and challenges. Effective ESC is based on relational, psychological, and physical theories \citep{rains2020support} and has been widely explored in artificial intelligence research \cite{liu2021ESconv, zhao-etal-2023-transesc}. With advancements in LLMs, these models have shown strong performance in ESC \cite{zheng-etal-2023-augesc, kang-etal-2024-large}. However, most LLM-based studies focus on immediate solutions without long-term support strategies. For example, while \citet{liu2021ESconv} defines ESC in three stages (Exploration $\rightarrow$ Comforting $\rightarrow$ Action), LLMs often struggle with smooth transitions, leading to strategy biases.

Motivated by the recent progress of reinforcement learning (RL) on LLM-based studies \cite{liDialogueActionTokens2024a,zhouArCHerTrainingLanguage2024a, wangImprovingMultistepReasoning2024a}, we propose that ESC tasks can be defined as a strategy-level MDP, therefore value-based RL can help mitigate the aforementioned challenges. Given the current seeker's utterance, emotion and conversational history, the LLM can be prompted to identify the long-term return of strategy, learn and produce the action value, and plan the optimal strategy. The determined strategy can then be prompted to another LLM to produce an improved response, guided by the strategy.

%Providing the current seeker's utterance and conversational history, the LLM is prompted to recognize the seeker's current emotion, determine its support strategy, and generate its own utterance. After the current conversational turn ends, the history is append with the emotion, strategy and utterances of the current turn, and the system returns back to the starting state. %Assuming the emotion and strategy lists are pre-defined and fixed, the emotion and strategy determinations have finite search space which facilitates the LLM to better reason and plan. Figure \ref{fig:paradigm} visualizes this paradigm.

\begin{figure}[t]
    \centering
    \includegraphics[width=\columnwidth]{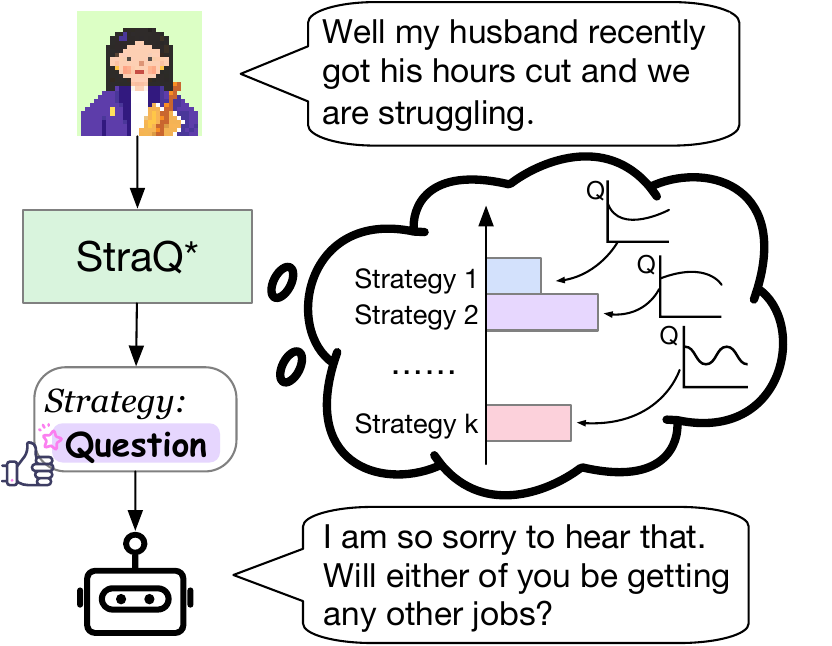}
    \caption{Paradigm of {\ModelName}. A plug-and-play LLM-based planner selects the optimal strategy from maximized $Q$, then steers the LLM to enhance the response.}
    \label{fig:p1}
\end{figure}
% long-term return optimization, and steers the conversational foundation model to have an enhanced response.

In this paper, we propose a new framework called strategic Q* ({\ModelName}), which converts LLM into a value-based strategic planner. We use deep Q-learning (DQN) on LLM to provide a strategic Q function, with the strategy as the textual action. We use the averaged logits of actions to denote the Q value, and update the LLM parameter by the famous Bellman equation. By this manner, we convert the next-token prediction to next-strategy prediction, bootstrapping the TD loss of strategies instead of the original cross-entropy loss. This Q-net is used as a plug-and-play strategic planner, along with the conversation LLM to produce the ultimate response. Our main contributions can be summarized as follows:\\
%\begin{itemize}
\noindent (1) We define the strategy-level MDP, and formulate the LLM architecture as a Q-function with textual input of state and strategy.\\
\noindent (2) We empirically verify that pretrained LLM can be finetuned by the Bellman Equation and converges to optimum returns, with the averaged logit of action tokens as the q-value.\\
\noindent (3) Substantial experiments on ESConv and EmpatheticDialogues indicate that {\ModelName} results in higher response quality and more reasonable planning of strategies.\\
\noindent (4) We design two reward mechanisms including imitation and distillation, with the former better at automatic metrics, while the latter better at human scoring and generalization.
%\end{itemize}

\section{Preliminary}

%\subsection{Q-Learning}

%\section{Preliminaries: Language Generation as a Reinforcement Learning Task}
%\label{sec:prelim}

\paragraph{Strategy-level MDP.}
The Markov decision process (MDP) is usually defined as a 5-tuple $(\mathcal{S}, \mathcal{A}, \mathcal{R}, \mathcal{T}, \gamma)$, where $\mathcal{S}$ is the state set, $\mathcal{A}$ is the action set, $\mathcal{R}$ is the reward set, $\gamma$ is the discounting factor of rewards, and $\mathcal{T}: \mathcal{S} \times \mathcal{A} \rightarrow \mathcal{S}$ is the state transition function.
In this work, we formalize the ESC task as a strategy-level MDP, with the action space $\mathcal{A} = \{ a \}$ as the set of possible strategies.
% a_{t} \in 
%, $pi(a|s)$ is the policy
% , \mu_0

\paragraph{Q-Learning.} In value-based RL, the goal is to learn the state-value function $V(s)$ or the state-action value function $Q(s, a)$, such that the determined action achieves the highest expected discounted cumulative reward:
\begin{align}
    a^{\star} = &\arg\max_{a} Q(s, a) \leftarrow \arg\max \sum_{t=0}^{\infty} \gamma^t r(s_t, a_t) \notag %\label{eq:maxQ} \\
    %\leftarrow &\arg\max R_t = \sum_{t=0}^{\infty} \gamma^t r(s_t, a_t) \notag
\end{align}
which is solved by the famous Bellman Equation: % Equation \ref{eq:maxQ}
\begin{align}
    Q^*(s, a) = r(s,a) + \gamma \max_{a'}Q^*(s', a')
\end{align}
in which the superscript $'$ indicates the next step. Instead of explicitly implementing the above equation, Deep Q-learning (DQN) approximates the maximization of the right-hand side with the deep value networks:
\begin{align}
    \mathcal{L}(\theta) = | r(s,a) + Q_{\phi}(s', a') - Q_{\theta}(s, a)|^2 \label{eq:dqn}  % V_{\phi}(s')
    %\mathcal{L} = |r + Q' - Q|^2
\end{align}
where $\mathcal{L}$ is the loss, $\theta$ and $\phi$ are parameters of the Q-net and the target Q-net, respectively. $\phi$ can be periodically synchronized from $\theta$. %  the action-averaged

\begin{figure}[t!]
    \centering
    \includegraphics[width=\columnwidth]{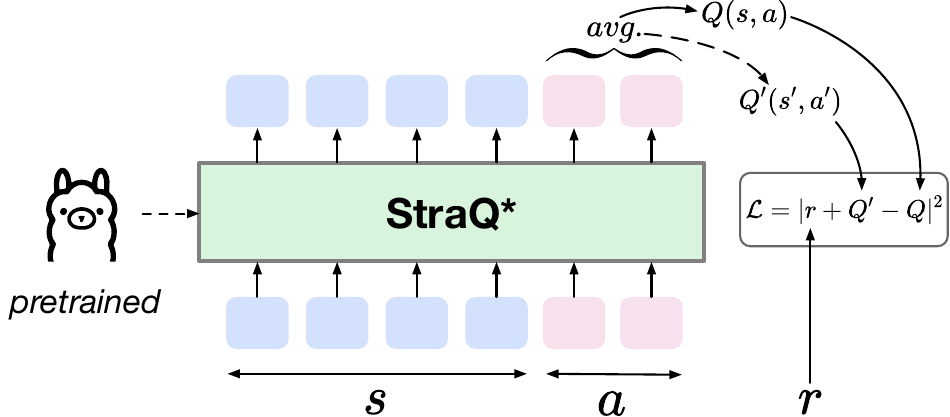}
    \caption{The training framework of {\ModelName}. Averaged log probability of action tokens is defined as $Q(s,a)$ which deduces the training loss from Bellman Equation.}
    \label{fig:p2}
\end{figure}

\section{Methodology}

\subsection{Task Definition}

%The problem of emotional-support conversation (ESC) can be characterized by an interleaved sequence of seeker and supporter utterance, 
%\begin{equation}
    %\{ u^k(t), u^p(t) \}_{t = 0, \dots, T}
    %\{ query, resp \}_{t \in 0:T}
%\end{equation}
%in which $T$ is the total number of conversation turn, $query$ and $response$ are utterances of seeker and supporter, respectively.

%$u$ denotes the utterance, $T$ is the total number of conversation turn, and $k$, $p$ represent seeker and supporter, respectively.

The problem of emotional-support conversation (ESC) can be characterized by an interleaved sequence of seeker $query$ and supporter $response$. To strengthen emotional-support performance, recent studies \cite{liu2021ESconv,rashkin-etal-2019-towards} enhance the data content by augmenting the set of support strategies $\mathcal{A}$ and seeker emotions $\mathcal{E}$. For each conversation session, the background $description$ is also annotated on the session-level. Such augmented ESC can then be described as
\begin{align}
    %& desc, \{u^k(t), e(t), a(t), u^p(t) \}_{t = 0, \dots, T} \label{eq:esc_seq} \\
    & desc, \{query(t), e(t), a(t), resp(t) \}_{0:T}  \label{eq:esc_seq} \\
    & a \in \mathcal{A}, \quad e \in \mathcal{E} \notag
\end{align}
in which $desc$ and $resp$ are the abbreviations of $description$ and $response$, and $T$ is the total number of conversation turns. At turn $t$, we denote the conversation history as
%During the conversation, the supporter agent needs to consider the current conversation background and the seeker's emotion $e$, choose the appropriate strategy $g$, and finally generate utterances $\{ u^p \}$ to obtain reasonable seeker's satisfaction.

\begin{align}
    h(t) = \{query(t), e(t), a(t), resp(t) \}_{0:T-1} 
\end{align}
Then the ESC sample at time $t$ can be alternatively expressed as $\{ h(t), query(t), e(t), a(t), resp(t) \}$.

%, which is consisting of the tripler (seeker utterance, strategy, supporter utterance).

%The emotional-support quality can be annotated from the utterance contents.

\subsection{System Variables}

%To develop our multi-turn emotional support conversation system, we adopt the concept of a finite state machine (FSM) to structure multi-turn emotional support dialogues. We define the dialogue sequence and the responses to be inferred across four stages: \textbf{history}, \textbf{state}, \textbf{rule}, and \textbf{answer}. Based on the current state of the dialogue seeker and historical input, the model determines the appropriate rule to follow for generating responses. In addition, the emotional responses of the model are guided both by the state and the selected rule. In the following, we provide detailed definitions and examples for each of these four components.

%Our {\ModelName} can be formally described as a 4-tuple $(\mathcal{S}, \mathcal{A}, \mathcal{R}, \mathcal{T})$, where: % , s_0

We define important system variables as follows:

%\begin{itemize}
\noindent $\bullet$ \textbf{State:} The state is a combination of description, emotion, history and query, \textit{i.e.}, $s = \{ desc, e, h, query \} \in \mathcal{S}$.\\
\noindent $\bullet$ \textbf{Action:} The conversational strategy, $a \in \mathcal{A}$.\\
\noindent $\bullet$ \textbf{Reward:} The reward $r_t$ can be viewed as the instantaneous satisfaction of the seeker, which can be either inferred from an annotated datasets, or generated by an off-the-shelf model evaluator (LLM-as-the-Judge).\\
%\item $\mathcal{C} = \{ desp \}$ is the set of job descriptions  % c \in\mathcal{C}
\noindent $\bullet$ $\mathcal{T}: \mathcal{S} \times \mathcal{A} \rightarrow \mathcal{S}$ is the transition function. After the $t$-th turn, $h$ is updated by appending the current $query(t)$ and $resp(t)$, the seeker reacts further with new $query(t+1)$ and $e(t+1)$, and the step is incremented by one.
%\item $s_0$ is the initial state. When $t=0$, $h = \emptyset$ and $s_0 = [u^k(t=0)]$
%\end{itemize}

\subsection{Implementation on Language Models}

\paragraph{LLM-based value function.} Our implementation starts from a pretrained LLM, with the parameter of $\theta$.  
We assume there is an instruction template with the placeholder of $s$, denoted by $\mathcal{I}(s)$. This instruction can be concatenated with $a$, $\mathcal{I} \oplus a$. Both state and state-action values can be obtained from the semantic understanding of LLM:
\begin{align}
    %V_{\theta}(s) &\leftarrow \text{LLM}_{\theta}(\mathcal{I}(s)) \\
    Q_{\theta}(s,a) &\leftarrowtail \text{LLM}_{\theta}(\mathcal{I}(s) \oplus a)
\end{align}
where $\leftarrowtail$ means to average the action logits.

% From this manner, the Q-net adopts the decoder architecture of the language model. 
% the Q-net are unified into a uniform architecture of the language model %, and the target net (as well as its parameter $phi$) in Equation \ref{eq:dqn} is omitted
\paragraph{Training a strategic value-function.} By replacing the Q-net in Equation \ref{eq:dqn} by the above expressions, we finetune the LLM by the Bellman Equation loss on last token logit. As in standard language modeling, the causal masking of the transformer allows us to perform Bellman updates on entire sequences in parallel. Figure \ref{fig:p2} exhibits this training framework.

We keep the setting of the target Q-net, which is the same LLM architecture, while its parameter $\phi$ is periodically synchronized from $\theta$.

\paragraph{Inference the optimal strategy.} Instead of decoding the next token, the finetuned LLM produces logits of available strategies, and the optimal strategy can be determined from the maximum logit
\begin{equation}
   a^{\star} \leftarrow \arg \max \text{LLM}(\mathcal{I}(s) \oplus a), a \in \mathcal{A}
\end{equation}

\begin{figure}
    \centering
    \includegraphics[width=\linewidth]{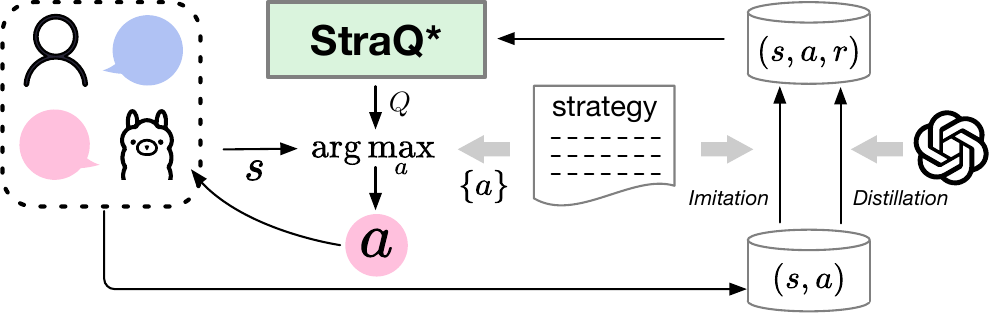}
    \caption{Diagram of our pipeline. Reward is annotated by the imitation or distillation strategy. The action is chosen from the strategy which maximizes the $Q$ value. }
    \label{fig:diagram}
\end{figure}

\paragraph{Instruction template.} We briefly exhibit our instruction $\mathcal{I}(s)$ here: 

\tcbset{
  colframe=black!75!white,
  colback=gray!5!white,
  boxrule=0.5pt,
  arc=2mm,
  left=1mm, right=1mm, top=1mm, bottom=1mm,
  fonttitle=\bfseries,
  before skip=5pt, after skip=5pt
}

\begin{tcolorbox}[title=Prompt Template]
\textbf{Description:} \{$desc$\} 
\ \textbf{\  User's emotion:} \{$e$\} \\
\textbf{History:} \{$h$\} \   
\textbf{\  Query:} \{$query$\} \\[2pt]
\textbf{Please select the best strategy:}
\vspace{-3pt}
\begin{itemize}[
    leftmargin=15pt,  
    labelwidth=10pt,     
    labelsep=5pt,         
    itemindent=0pt,
    align=left,
    itemsep=1pt, topsep=2pt, parsep=0pt
]
  \item[\textbf{(1)}] \{$strategy 1$\} \textbf{\ (2)} $\cdots$ \textbf{\ (K)} \{$strategy K$\}
  % \item[$\cdots$]
  % \item[(K)] \{strategy K\}
\end{itemize}
%\textbf{Answer:} %(a)
\end{tcolorbox}
\noindent with the full version in Appendix \ref{sec:detailed_prompt}. To further strengthen the understanding capability of LLM on the strategy selection, we formulate $\mathcal{I}$ as a multi-choice question (MCQ), instead of a plain question, forcing the LLM to choose one of the option numbers. Accordingly, the action set becomes the set of possible strategy index $a \in \mathcal{A} := \{1, 2, \cdots, K \}$ where $K$ is the total number of strategies.

\subsection{Reward Definitions}

Choice of rewards may be crucial especially when the sampling is constrained by an offline dataset. In this paper, we study two reward mechanisms: \\
%\begin{itemize}
\noindent (1) Distillation: for each ($s$, $a$) pair from the dataset, we let a strong-basis LLM (\textit{e.g.}, GPT-4) to provide a judge score from 0 to 5 (The detailed judge prompt is in Appendix \ref{sec:detailed_prompt}). Since this manner distills the knowledge from a teacher model, we call this variant as \textbf{{\ModelName}-distill}.\\
\noindent (2) Imitation: we consider each ($s$, $a$) pair from the dataset is an expert demonstration, therefore, always assigned with $r$ of $+1$. To amplify the distribution, we randomly sample a different $a$ and assign with $r$ of $-1$. The positive-negative ratio is 1:1. Since this manner imitates the positive samples directly, this variant is called \textbf{{\ModelName}-imit}.
%\end{itemize}

Figure \ref{fig:diagram} shows the entire pipeline of {\ModelName}.

%We expect {\ModelName}-imit might have better automatic metrics which are based on similarity while {\ModelName}-distill would perform better on human evaluation.

\begin{figure}[t]
    \centering
    \includegraphics[width=0.9\columnwidth]{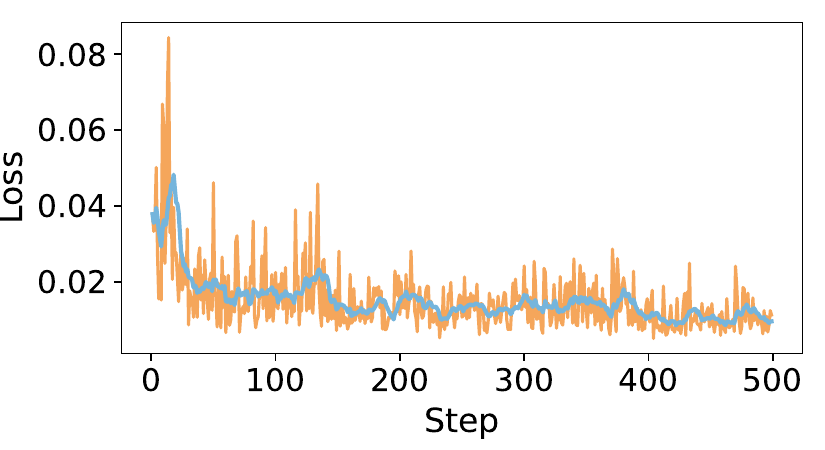}
    \caption{Training loss curve of {\ModelName}. }
    \label{fig:loss}
\end{figure}

\begin{table}[h!]
\centering
\small
%\setlength{\tabcolsep}{8pt} % 调整列间距
%\begin{tabular}{>{\centering\arraybackslash}m{0.15\textwidth}|m{0.05\textwidth}|m{0.65\textwidth}}
\begin{tabular}{l|c|c}
\toprule
\multicolumn{1}{c|}{\textbf{Strategies}} & \multicolumn{1}{c|}{\textbf{Abbr.}} &\multicolumn{1}{c}{\textbf{Stage}}  \\ 
\toprule
Question & \bf Que.& I \\ 
%\hline
Restatement or Paraphrasing &\bf Res.\& Par.& I \\ 
%\hline
Reflection of Feelings &\bf Ref.& II \\ 
%\hline
Self-disclosure &\bf Self-Dis.& II \\ 
%\hline
Affirmation and Reassurance &\bf Aff.\& Rea.& III \\
%\hline
Providing Suggestions &\bf Pro.& III \\ 
%\hline
Information & \bf Inf.& III \\ 
%\hline
Others &\bf Others& - \\ 
\bottomrule
\end{tabular}
\caption{Strategy names, abbreviations and stages.}
\label{tab:strategies}
\end{table}

\section{Experiment}

\subsection{Setting}

\paragraph{Implementation.} Llama3.2-1B-instruct \cite{llama3modelcard} is employed as the base model. Training is conducted on OpenRLHF \cite{hu2024openrlhf}, with the learning rate of $5.0e-6$, window length of 2048, batch size of 64, and epoch of 4. The target network update frequency is set to 10, the replay buffer size is 12,000, and $\gamma = 0.85$. %Experiments are conducted on 8 NVIDIA A100 80GB PCIe GPUs. 

\paragraph{Datasets.} Training of {\ModelName} requires the annotation of strategies. We use ESConv \cite{liu2021ESconv} as the training set and also the in-domain (ID) test. ESConv provides $K=8$ strategies, which belong to three ESC stages: \textit{Exploration (I), Comforting (II)} and \textit{Action (III)}. Table \ref{tab:strategies} shows their full names, abbreviations and corresponding stages.

Furthermore, EmpatheticDialogues \cite{rashkin-etal-2019-towards} is employed as the out-of-domain (OOD) evaluation, since EmpatheticDialogues does not have the strategy annotation. For the ID test, both strategy-related and response-related results can be provided. For the OOD test, only zero-shot response-related results are provided. Appendix \ref{sec:dataset} provides a more detailed introduction of ESConv and EmpatheticDialogues.
%we evaluate the performance of the model finetuned on ESConv, with 

\subsection{Evaluation Methods}
\label{sec:evaluation}

%\subsection{Evaluation on Strategy Planning}
%\paragraph{Proficiency.}

\paragraph{Automatic Metrics.}
To evaluate the quality of strategy determination, we refer the evaluation methods proposed by \citet{kang-etal-2024-large}, which uses \textbf{proficiency} $\mathcal{Q}$ based on macro-F1, and \textbf{preference bias} $\mathcal{B}$ based on Bradley-Terry model~\citep{bradley1952btmodel}. Smaller $\mathcal{B}$ means less bias, therefore is better. We also include the strategy prediction accuracy (Acc). For response quality, we utilize the famous Bleu-2 (B-2), Rouge-L (R-L), Distinct-2 (D-2) and CIDEr, calculating from the similarity with the ground truth response.

\paragraph{Human Scoring.}
Similar with \citet{kang-etal-2024-large}, we annotate with the dimensions of \textit{Acceptance}, \textit{Effectiveness}, \textit{Sensitivity}, \textit{Fluency}, and \textit{Emotion}, and the ultimate purpose, seeker's \textit{Satisfaction}. %Scores are averaged over results of ESConv and EmpatheticDialogues.

%\paragraph{LLM-as-a-Judge.}
% \textcolor{red}{@zhaoyue}
%We used GPT-4o to compare different responses and provide the win-tie-lose suggestion. The evaluation prompt is obtained from \citet{madani2024steeringconversationallargelanguage} with details in Table \ref{tab:prompt_comparison_score}.

\paragraph{Baselines.} We consider the following baselines:

%\begin{itemize}
\noindent (1) Direct: directly inference the LLM.\\ % , with the same context
\noindent (2) Direct-Refine: the model immediately revises its response within the same utterance to incorporate emotional support considerations.\\
\noindent (3) Self-Refine \citep{Madaan2023SelfRefine}: the model considers the emotional support, generates a feedback from the initial response, then refines the response based on the feedback.\\
\noindent (4) CoT \citep{wei2022chain}: steered by the chain-of-thought prompt, the model first identifies \textit{emotion}, then generates \textit{strategy}, and finally \textit{response}.\\
\noindent (5) FSM \cite{wangFSMFiniteState2024}: the finite state machine with finite sets of states and state-transitions triggered by inputs, and associated discrete actions.
%for analyzing discrete-state deterministic systems.
    %\item SFT: finetuning on the training sets, including the strategy and emotion labels.
%\end{itemize}

%We compare performances of methods on the basis of both the original Llama3-8B-instruct and its finetuned version\footnote{Conduct SFT of Llama3-8B-instruct on the training set.}, respectively.

%Concatenates history, state, and response into a single output using special characters. The history is used as input, and the model is fine-tuned on the aforementioned data to perform multi-stage inference.

\begin{table}[t!]
\renewcommand{\arraystretch}{1.11}
\centering
\small
\resizebox{\columnwidth}{!}{
\begin{tabular}{l ccccc}
    \toprule
    Methods & Acc $\uparrow$ & $\mathcal{Q}$ $\uparrow$ & $\mathcal{B}$ $\downarrow$ & B-2 $\uparrow$ & R-L $\uparrow$ \\
    \midrule
    \textit{LLaMA3-8B-Instruct} \\ 
    \midrule
    Direct & 11.80 & 10.26 & 1.61 & 3.47 & 10.64 \\ % (\textit{0-shot})
    + Direct-Refine & 17.08 & 11.07 & 1.27 & 3.10 & 6.13 \\
    + Self-Refine & 17.58 & 13.61 & 1.92 & 3.34 & 9.71 \\
    + CoT & 15.32 & 10.38 & 1.69 & 3.16 & 10.50 \\
    + FSM & 17.37 & 11.15 & 0.81 & \textbf{4.12} & \underline{11.83} \\
    \textbf{+ 1B {\ModelName}-distill (ours)} & \underline{41.22} & \underline{38.95} & \textbf{0.57} & \underline{3.89} & 11.80 \\
    \textbf{+ 1B {\ModelName}-imit (ours)} & \textbf{46.83} & \textbf{43.15} & \underline{0.80} & \underline{3.89} & \textbf{12.84} \\
    \midrule
    \textit{LLaMA3-8B-Instruct + SFT} \\ 
    \midrule
    Direct & 32.43 & 21.29 & 1.28  & 6.97 &  16.59  \\
    + CoT & 30.80 & 17.70 & 1.35  & 6.51 &  15.00  \\
    + FSM & 28.83 & 18.36 & 1.32  & \underline{7.57} &  \textbf{17.42}  \\
    \textbf{+ 1B {\ModelName}-distill (ours)} & \underline{41.22} & \underline{38.95} & \textbf{0.57} & 7.01 & 16.93 \\
    \textbf{+ 1B {\ModelName}-imit (ours)} & \textbf{46.83} & \textbf{43.15} & \underline{0.80} & \textbf{7.63} & \underline{17.30} \\
    \bottomrule
\end{tabular}}
\caption{ID Results of automatic metrics including Acc, $\mathcal{Q}$, $\mathcal{B}$, Bleu-2 (B-2) and Rouge-L (R-L) on the testset of ESConv. The best results of each LLMs are \textbf{bolded} and the second best are \underline{underlined}. \\ % The best result across all base models is marked blue.
%  A single strategy planner is employed to predict strategies and provides them to each LLM.
%  and the second best are \underline{underlined}
%$^{\ast}$: results are obtained directly from \citet{kang-etal-2024-large}.
}
\label{tab:methodology_results}
\end{table}

\begin{table}[htbp!]
\renewcommand{\arraystretch}{1.11}
\centering
\small
\resizebox{0.98\columnwidth}{!}{
\begin{tabular}{l cccc}
    \toprule
    \textbf{Methods} & B-2 & R-L & Dist-2 & CIDEr \\
    \midrule
    Direct & 3.09 & 9.91 & 25.23 & 1.60  \\
    %Direct & 4.10 & 10.88 & 47.56 & 6.84  \\
    + CoT & 2.91 & 9.79 & 32.65 & 1.37 \\
    %+ CoT & 5.05 & 14.06 & 27.30 & 12.01 \\
    + FSM & 3.33 & 10.80 & 33.37 & 2.96 \\
    + \textbf{1B {\ModelName}-distill (ours)} & \bf 4.49 & \bf 12.93 & \underline{46.53} & \bf 8.36 \\
    + \textbf{1B {\ModelName}-imit (ours)} & \underline{4.27} & \underline{12.66} & \bf 46.80 & \underline{8.11} \\
    \bottomrule
\end{tabular}
}
\caption{OOD finetuned results of Bleu-2 (B-2) and Rouge-L (R-L) on EmpatheticDialogues. The best results of each LLMs are \textbf{bolded} and the second best are \underline{underlined}. %Numbers in blue mean comparable better results across all base modes.\\
%  A single strategy planner is employed to predict strategies and provides them to each LLM.
%$^{\ast}$: results are obtained directly from \citet{kang-etal-2024-large}.
}
\label{tab:auto_empatheticdialogues}
\end{table}

\begin{table*}[htbp!]
\centering
\resizebox{\textwidth}{!}{%
\begin{tabular}{l|ccccccc}
    \toprule
    \multicolumn{1}{c|}{\multirow{2}[4]{*}{Method}} & \multicolumn{7}{c}{Human Annotation} \\
%\cmidrule{1-2}  
\cmidrule{2-8}    
 & Fluency  & Emotion & Acceptance & Effectiveness & Sensitivity & Alignment & Satisfaction \\ % Interesting % Total % $\uparrow$
    \toprule   
    Original dataset & 3.51 & 3.61 & 3.40 & 3.10 & 3.50 & 3.20 & 3.30 \\
    \midrule
    %\multicolumn{1}{c|}{} & Prompt+Qwen2 & 72B  & - & - & - & - & - & - & - \\
    %\multicolumn{1}{c|}{} & Prompt+Mistral & 8x7B  & - & - & - & - & - & - & - \\
    %\multicolumn{1}{c|}{\multirow{11}[2]{*}{8B}} 
    Llama3-8B-Instruct  & 2.95 & 3.00 & 2.60 & 2.40 & 2.70 & 2.70 & 2.60 \\
    \;+ Direct-Refine & 3.09 & 3.09 & 2.73 & 2.91 & 2.91 & 2.82 & 2.84 \\
    \;+ Self-Refine & 3.10 & 3.15 & 2.80 & 2.70 & 2.90 & 2.80 & 2.80 \\
    \;+ CoT  & 3.08 & 3.08 & 2.83 & 2.67 & 3.00 & 2.83 & 2.83 \\
    \;+ FSM  & 3.30 & 3.35 & 2.90 & 2.90 & 3.00 & 2.90 & 2.93 \\
    \midrule
    \;Llama3-8B-Instruct+ {\vanillaFT} & 3.15 & 3.40 & 2.70 & 2.70 & 2.90 & 3.30 & 2.90 \\
    \;+ CoT  & \bf 3.67 & 3.61 & 3.22 & 3.67 & 3.56 & 3.35 & 3.45 \\
    \bf \;+ {\ModelName}-distill (ours) & 3.52 & \bf 3.65 & \bf 3.59 & \bf 3.73 & \bf 3.71 & \bf 3.62 & \bf 3.66 \\
    \bf \;+ {\ModelName}-imit (ours) & 3.42 & 3.25 & 3.23 & 3.07 & 3.10 & 3.21 & 3.13 \\
    \bottomrule
    \end{tabular}%
}
\caption{Averaged Human evaluation of response quality on ESConv and EmpatheticDialogues.}
\label{tab:response_quaility}
% Value with bold indicates the best results while value with underline indicates the second best results. We do not assess non-LLM methods because they are not open-sourced.
%${\ast}$: Value with bold indicates the best results while value with underline indicates the second best results. We do not assess non-LLM methods because they are not open-sourced.
\end{table*}

\subsection{Results}

\paragraph{Training Curves.} Figure \ref{fig:loss} shows the training loss curve of \ModelName for 500 steps (approximately 3 epochs). Although the loss initially fluctuates significantly, it adapts to the new training paradigm, and finally tends to be stable.

\paragraph{Automatic Evaluations.} Table \ref{tab:methodology_results} presents the automatic metrics on the ID evaluation, with the basis of either the original LLM, or the specifically finetuned version. Compared to baselines, \ModelName generally achieves higher strategy accuracy, lower bias, and higher similarity to the ground truth responses. Furthermore, \ModelName-imit performs better than \ModelName-distill on this setting, suggesting that the imitation-version of rewards result in better ID performance.
%shows higher Acc due to dataset-derived rewards, this does not imply superiority over \ModelName-distill, as discussed later. 

In Table \ref{tab:auto_empatheticdialogues}, we further compare the OOD results of the models finetuned by ESConv, with the strategy lists inferred from ESConv. Results suggest that \ModelName demonstrates strong generalization than these baselines. Specifically, \ModelName-distill surpasses \ModelName-imit this time, indicating the distilled knowledge from the teacher model is more general than simply imitating a limited dataset.

%. Details of per-strategy results are shown in the appendix.

% \begin{table}[htbp]
%     \centering
%     \begin{tabular}{l|l|l|l|l|l}
%         \hline
%         score & 1 & 2 & 3 & 4 & 5 \\
%         \hline
%         GPT\-4 & 0.05\% & 0.72\% & 36.04\% & 58.53\% & 4.66\% \\
%         \hline
%         Acc & 0\% & 0\% & 6.92\% & 54.96\% & 90.77\% \\
%         \hline
%     \end{tabular}
% \end{table}

\paragraph{Human Evaluation.} The results of the crowdsourcing evaluation shown in Table \ref{tab:response_quaility} indicate that \ModelName-distill outperforms the baseline methods in various metrics, such as Fluency, Emotion, and Satisfaction. It also performs better than the replies in the source data. Conversely, \ModelName-imit is slightly lower than the source data in performance. Using the GPT-4 score as a reward, {\ModelName} can determine strategies more from the aspect of performance optimization, not simply imitating the demonstration.

%This is because the reward signal of \ModelName-distill is the GPT-4 score, which prevents the generation of obviously poor responses present in the dataset.

%\paragraph{Comparison results.} Table \ref{tab:gpteval_results}

\begin{figure*}[htbp!]
    \centering
    \includegraphics[width=0.48\linewidth]{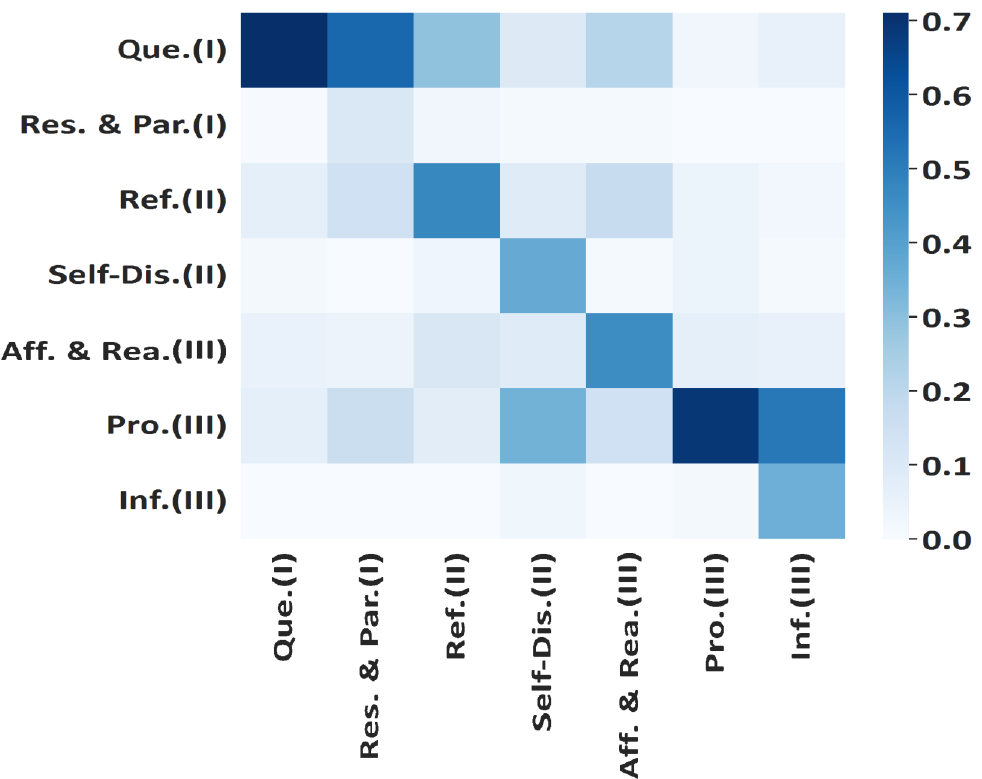}
    \hspace{0.1in}
    \includegraphics[width=0.48\linewidth]{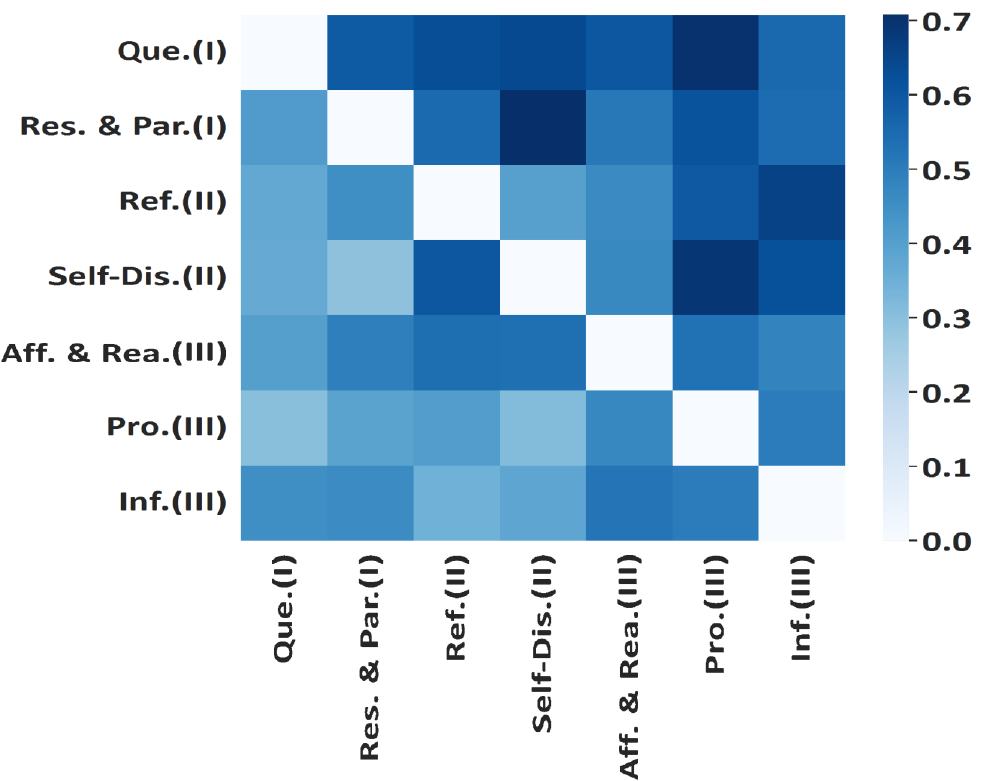}
    %\vspace{-3.5mm}
    \caption{Distribution of strategy determined by {\ModelName}. Strategies are labeled with the stage index (I, II, III) which represents the general scenario Exploration $\rightarrow$ Comforting $\rightarrow$ Action in ESC. Left: the confusion matrix (acted strategy (row) VS ground truth strategy (column)). Right: the transition matrix (acted strategy (row) VS the next-acted strategy (column)).\\
    }
    \label{fig:strategy_matrix}
\end{figure*}

%\subsection{Ablation Study}

\paragraph{Ablation Study.} Two ablations are studied:

%\begin{itemize}
\noindent (1) \textit{w/ value head}: append the model with a classification head which produces the score logit. \\
    %, which is the usual implementation of the reward model in RLHF. 
    %is connected to a fully-connected network, and the output layer has only one neuron serving as the strategy value.
\noindent (2) \textit{auto-regressive}: keep the cross-entropy loss with the ground truth action as the target text.\\
%\end{itemize}
In more detail, \textit{w/ value head} is the usual solution for a reward model in RLHF, while \textit{auto-regressive} can be viewed as a standard fine-tuning solution for the strategic planner. Table \ref{tab:ablation} shows that \ModelName outperforms both of them in various automatic metrics, indicating our methodology can better align with the strategy semantics and more accurately capture the strategic value.

%\subsection{Sensitivity Analysis}

\paragraph{Sensitivity Analysis.} Figure \ref{fig:sensitivity} shows the ID performance evolutions on different $\gamma$ choices. Smaller $\gamma$ means we are more focused on the transient performance and relatively neglect the long-term value. Results show that the optimal accuracy of strategy happens on $\gamma = 0.9$, while the best response-related metrics correspond to $\gamma = 0.85$. Because B-2 and R-L are similarity-based, the current reward is more relative to them than future rewards. Therefore, this observation is reasonable. %  also

%shows when $\gamma$ is within a certain range (0.8 to 0.9), its impact on the final performance is relatively small. However, when $\gamma$ is 0.99, the performance of the model decreases significantly. In addition, as the model size increases from 1B to 8B, the model performance also rises slowly. Nevertheless, considering the comprehensive factors of inference speed and computing resources, a 1B model is already enough.
%size: bars of instruct B-2 w r t 1B 3B 8B

\subsection{Discussions}

\begin{figure}[ht]
    \centering
    \includegraphics[width=0.45\linewidth]{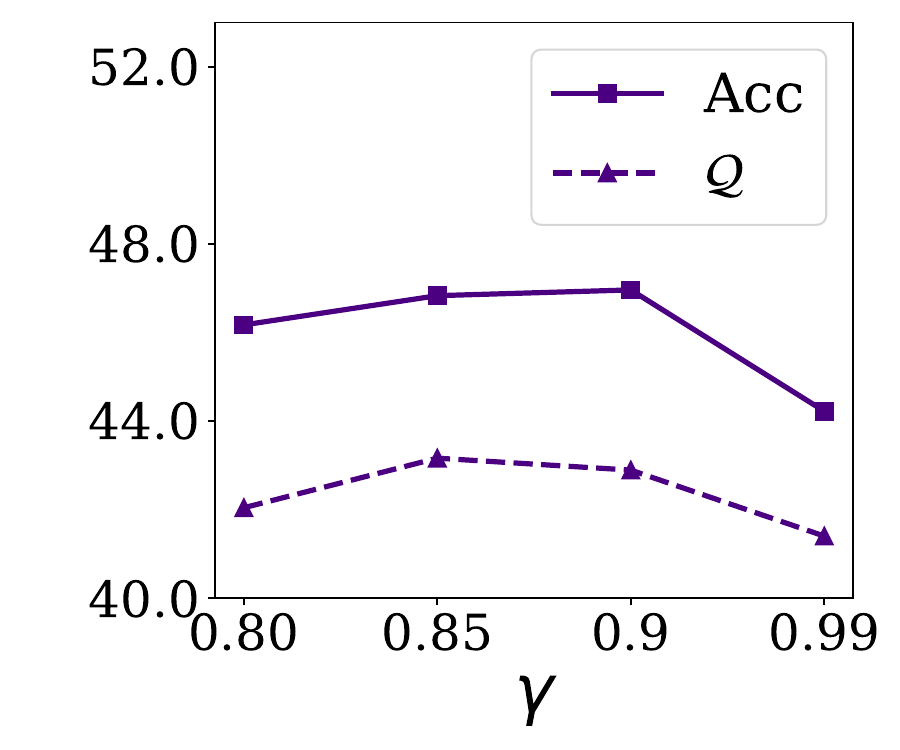}
    \hspace{0.1in}
    \includegraphics[width=0.45\linewidth]{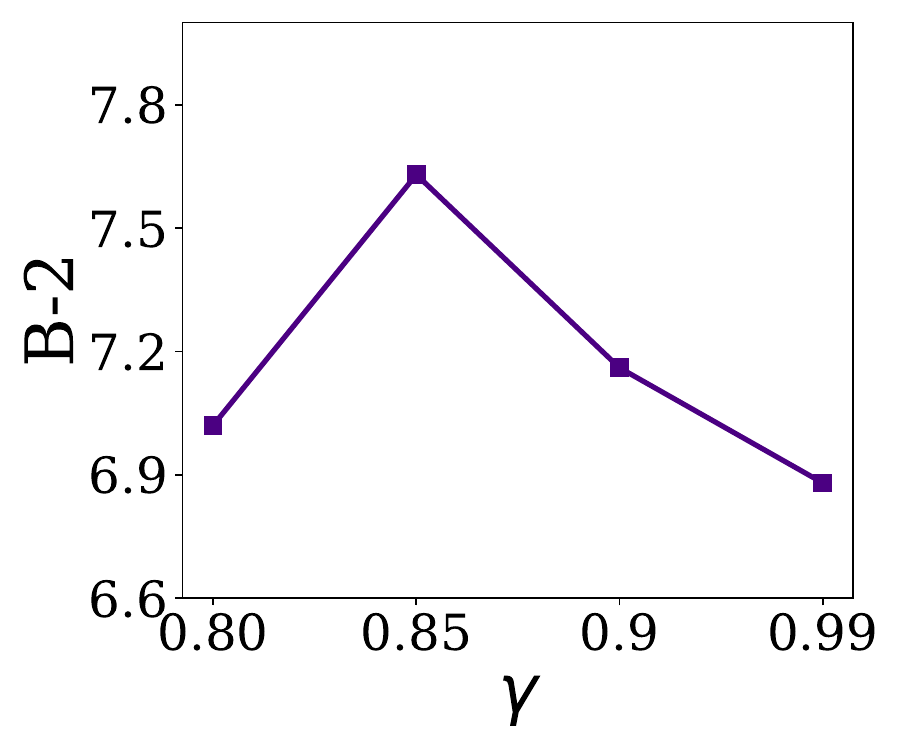}
    %\vspace{-3.5mm}
    \includegraphics[width=0.45\linewidth]{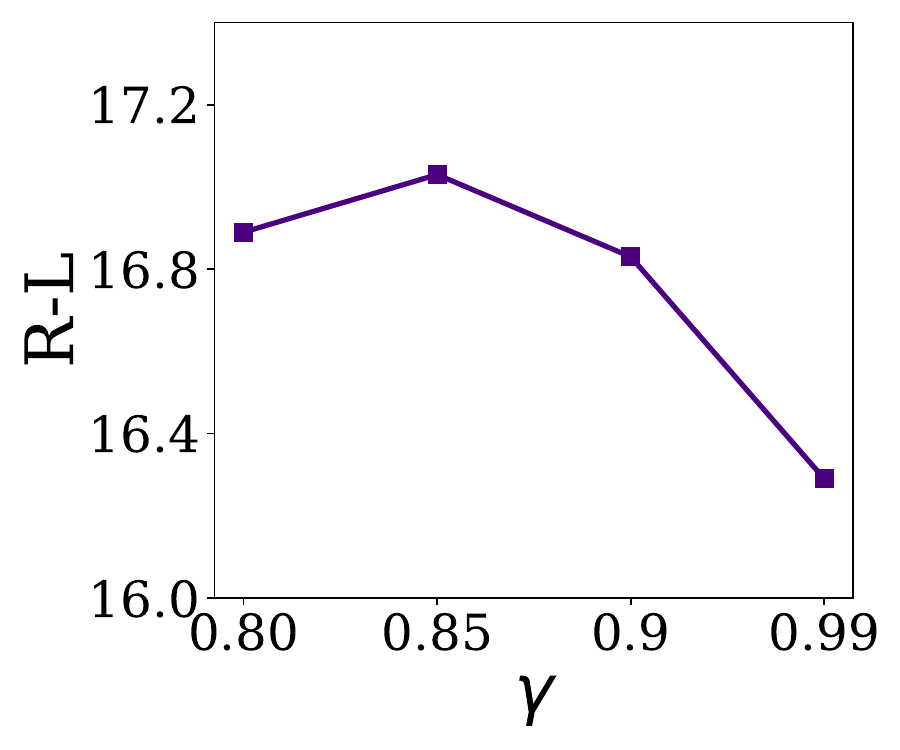}
    \hspace{0.1in}
    \includegraphics[width=0.45\linewidth]{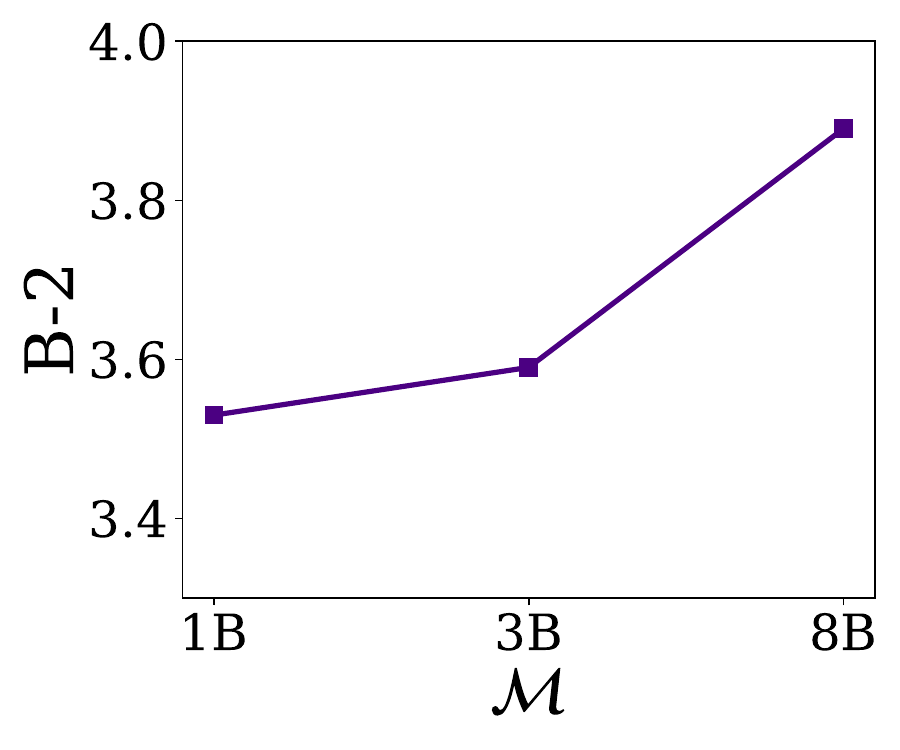}
    \caption{Sensitivity plots of {\ModelName}-imit on different $\gamma$ and model sizes. Metrics include Acc, $\mathcal{Q}$, B2, and R-L.} %  The results for $\gamma$ come from {\ModelName}-imit, while the results for model sizes are from {\ModelName}-distill.
    \label{fig:sensitivity}
\end{figure}

\paragraph{Scalability and application.} Figure \ref{fig:sensitivity} (bottom-right) also compares the B-2 results on different model sizes. As the model becomes larger, the performance also increases, indicating {\ModelName} can have good scalability. However, larger models result in higher computation overhead and slower speed, which may hinder the practical application of {\ModelName}. Therefore, in the formal application, we still adhere to the 1B choice, employing it as a lightweight planner.

From previous results, we utilize {\ModelName}-distill in the actual application to have better generalization and better alignment with human knowledge.

\paragraph{Returns of Strategies.} Table \ref{tab:reward_andvalue} further analyzes two important indicators of value-based RL, the averaged rewards and values. In this analysis, the rewards are provided by GPT-4. {\ModelName} achieve both higher <reward> and <value> than direct inference of the base model, as well as the annotation of the original dataset. This result shows that {\ModelName} statistically obtains higher returns, which is the primary purpose of Q-Learning.

%a statistical analysis of key performance indicators in reinforcement learning training. The <reward> represents GPT-4's evaluation of generated replies on the test set, while the <value> reflects the model's confidence in its decisions. Models trained with \ModelName exhibit high <value>s, and \ModelName-distill, leveraging GPT-4 scores as rewards, achieves a higher <reward>.

\begin{table}[h!]
\small
\begin{center}
\begin{tabular}{lcc}
\toprule
\textbf{Method} & <reward> & <value> \\
\hline
Original dataset   & 3.01  & 252.09  \\
Llama3-8B-Instruct  & 3.66  & 346.31 \\
\bf {\ModelName}-distill (ours) & \textbf{3.99}  & 424.78 \\
\bf {\ModelName}-imit (ours) & 3.72  & \textbf{445.95} \\
\bottomrule
\end{tabular}
\end{center}
\caption{Average reward (eval by GPT-4) of strategy determination on the test set of ESConv.}
\label{tab:reward_andvalue}
\end{table}

\begin{table}[t!]
\renewcommand{\arraystretch}{1.11}
\centering
\small
\resizebox{0.98\columnwidth}{!}{
\begin{tabular}{l ccccc}
    \toprule
    \textbf{Method} & Acc $\uparrow$ & $\mathcal{Q}$ $\uparrow$ & $\mathcal{B}$ $\downarrow$ & B-2 $\uparrow$ & R-L $\uparrow$ \\
    \midrule
    %LLaMA3-8B-Instruct + SFT \\ 
    w/ value head & 19.81 & 11.40 & 1.66 & 6.74 & 15.99  \\
    auto-regressive & 46.22 & 43.01 & \bf 0.69 & 7.25 & 16.48 \\
    %wo/ SFT & 5.05 & 14.06 & 27.30 & 12.01 \\
    \bf {\ModelName}-imit  & \bf 46.83 & \bf 43.15 & 0.80 & \bf 7.63 & \bf 17.03 \\
    \bottomrule
\end{tabular}}
\caption{Ablation study of {\ModelName}-imit on ESConv. %  of finetuned results
}
\label{tab:ablation}
\end{table}

\paragraph{Strategy Prediction and Transitions.} Figure \ref{fig:strategy_matrix} (Left) exhibits the confusion matrix of strategies,  with the rows representing the prediction, and the columns representing the ground truth. Results show that most occurrences happen on the diagonal grids, verifying the prediction accuracy.

Figure \ref{fig:strategy_matrix} (Right) visualizes the transition matrix. A grid $(i, j)$ means the strategy $i$ to the strategy $j$, where the strategies are sorted from their ESC stages (from I to III) for both rows and columns. Therefore, transitions from an early stage to a later stage should occur on the upper-triangle region of the transition matrix. The results in the figure validate this proposition.

%The confusion matrix shown in Figure \ref{fig:strategy_matrix} indicates that for most strategies, the color is the darkest along the main diagonal. In the transition matrix, the color in the upper right triangular area is significantly darker than that in the lower left area, which is consistent with the three-stage theory in ESC.

\begin{table*}[htbp]
\centering
\small
\resizebox{\textwidth}{!}{
\begin{tabular}{l cp{10.5cm} c c}
      \toprule[1.5pt]
        \textbf{Characters} & \textbf{Strategy}& \textbf{Response} & \textbf{Reward} & \textbf{Value}\\ 
        \midrule
        \textit{supporter}& - &\textit{Hi! Hope you are doing well?} & - & -\\
        % \midrule
        \textit{seeker}& - & \textit{Hi, yes am doing well}  & - & -\\
        % \midrule
        \textit{supporter}& - & \textit{How can I help you today?} & - & -\\
        % \midrule
        \textit{seeker}& - & \textit{I had a drinking challenge which has been affecting my marriage for quit sometime now } & - & -\\
        \midrule
        \textit{supporter} & \makecell[t]{ \textit{Res.\& Par.} \\(\textit{I})} & \textit{So as far as I understand that you had issues with your wife due to your drinking. Am I right?} & 3 & 7.6\\
        Direct & \makecell[t]{Res.\& Par. \\(I)} & So, it sounds like there's been a bit of tension in your marriage lately due to your drinking habits, and it even led to a confrontation that's caused a rift between you and your wife. Can you tell me more about what's been going on with your drinking? & 4 & 8.0\\
        \textbf{{\ModelName}-distill} & \makecell[t]{\textbf{Res.\& Par.}\\ (\textbf{I})} & \textbf{It sounds like your drinking habits have been causing difficulties in your marriage for a while.} & 3 & \textbf{10.1}\\
        \midrule
        \textit{seeker}& - & \textit{Yes that is true} & - & -\\
        \midrule
        \textit{supporter} & \makecell[t]{\textit{Aff.\& Rea.}\\ (\textit{II})} & \textit{That is really a serious problem. I know it must be very tough time for you. Hope you are trying your best to save your marriage.} & 2 & 5.4\\

        Direct & \makecell[t]{Res.\& Par. \\(I)} & Just to make sure I understand correctly, it sounds like there was a confrontation with your wife about your drinking habits, and now you two aren't speaking. Is that a fair summary of what happened? & 3 & 4.7\\
        \textbf{{\ModelName}-distill} & \makecell[t]{\textbf{Ref.}\\ (\textbf{II})} & \textbf{That must be really difficult for you. I can imagine how much stress and worry this situation has brought into your life.} & 4 & \textbf{8.3}\\
        \midrule
        \textit{seeker}& - & \textit{Yes I have really reduced my drinking to ensure I save my marriage} & -  & -\\
        \midrule
        \textit{supporter}& \makecell[t]{\textit{Ref}. \\(\textit{II})} & \textit{That's is a great initiative. I know it is tough for you but not impossible. I can feel how hard you are trying t save your relationship. I appreciate you for that.} & 4 & 4.0\\

        Direct & \makecell[t]{Res.\& Par. \\(I)} & Just to make sure I understand correctly, you've been trying to cut down on your drinking to address your wife's concerns, but despite your efforts, the issue still seems to be causing tension in your marriage, and now you're not talking to each other. Is that a correct summary? & 2  & 2.0\\
        \textbf{{\ModelName}-distill} & \makecell[t]{\textbf{Aff.\& Rea.} \\(\textbf{III})} & \textbf{That’s a great step forward! It takes a lot of strength and commitment to make such a change, and you should be proud of yourself.} & 5 & \textbf{5.0 }\\

        \bottomrule
\end{tabular}
}
\caption{Case of {\ModelName} in ESConv.}
\label{tab:case}
\end{table*}

\paragraph{Detailed results of Strategies.} Strategies' popularities and occurrences may differ in nature. For example, straightforward strategies like \textit{"Question"} and \textit{"Providing Suggestions"} may be more convenient to learn and apply. A reasonable strategic planner should reflect this frequency difference, but also avoid significant bias (\textit{e.g.}, always determines the most frequent strategy). To further study the strategy-related performance, we further exhibit the per-strategy results in Table \ref{tab:detail_b_q}, with the comparison between {\ModelName}-distill and SFT. One can observe that {\ModelName} model is more balanced across different strategies, and SFT even has zero predictions on the strategy of "Information".

%The performance of the model often differs among various strategies. For humans, straightforward strategies like \textit{"Question"} and \textit{"Providing Suggestions"} may be more convenient to learn and apply. As a result, these strategies also appear relatively more frequently in the source data. Table \ref{tab:detail_b_q} shows that for both the SFT and \ModelName models, the above-mentioned strategies also have higher values of Acc, $\mathcal{Q}$, and R-L value. Moreover, compared with the SFT model, in addition to having higher indicators such as the average Acc, the \ModelName model is more balanced among different strategies and avoids the occurrence of zero values.

\begin{table}[t] 
  \centering
  \caption{Per-strategy automatic metrics on ESConv.}
  \resizebox{\columnwidth}{!}{
    \begin{tabular}{c|l|ccccccc}
    \toprule
     & \textbf{Strategy} & Acc $\uparrow$   & $\mathcal{Q}$ $\uparrow$      & $\mathcal{B}$  $\downarrow$    & B-2 $\uparrow$  & R-L $\uparrow$  & Dist-2   & CIDEr\\
    \midrule
    \multirow{8}[2]{*}{\rotatebox{90}{SFT}} & Que.  & 57.52 & \textbf{48.24}  & 1.60  & \textbf{9.37}  & \textbf{21.88}  & 64.42 & \textbf{34.22} \\
          & Res.\& Par. & 18.52 & 20.55  & 0.84  & 7.96  & 16.99  & 77.38 & 21.12 \\
          & Ref.  & 1.57  & 2.92  & 0.08  & 5.74  & 14.90  & 73.44 & 11.72 \\
          & Self-Dis. & 2.36  & 4.38  & \textbf{0.06}  & 4.99  & 12.30  & 76.71 & 7.85 \\
          & Aff.\& Rea. & 20.09 & 22.45  & 0.83  & 5.94  & 15.17  & 70.72 & 13.20 \\
          & Pro.  & \textbf{75.22} & 40.95  & 4.12  & 5.99  & 14.26  & 71.26 & 11.04 \\
          & Inf.  & \textcolor{red}{0.00}   & \textcolor{red}{0.00}  & \textcolor{red}{0.00}  & 5.93  & 12.24  & \textbf{78.54} & 12.95 \\
          & Others & 23.20  & 30.77  & 0.46  & 8.38  & 18.21  & 74.08 & 27.78 \\
    \midrule
    \multirow{8}[2]{*}{\rotatebox{90}{\ModelName-distill}} & Que.  & \textbf{71.07} & \textbf{60.59}  & 2.43  & \textbf{9.48}  & \textbf{22.05}  & 64.44 & \textbf{33.50} \\
          & Res.\& Par. & 8.97  & 16.67  & \textbf{0.16}  & 8.45  & 17.30  & 79.05 & 21.60 \\
          & Ref.  & 40.88 & 38.69  & 0.43  & 5.18  & 13.62  & 75.95 & 9.78 \\
          & Self-Dis. & 29.85 & 41.75  & 0.46  & 4.90  & 12.70  & 76.00    & 6.01 \\
          & Aff.\& Rea. & 36.63 & 42.11  & 0.47  & 6.30  & 15.46  & 70.35 & 13.66 \\
          & Pro.  & 69.08 & 56.13  & 0.60  & 5.95  & 14.03  & 70.51 & 10.62 \\
          & Inf.  & 26.67 & 40.14  & 0.47  & 5.33  & 13.21  & \textbf{79.94} & 8.11 \\
          & Others & 45.93 & 45.95  & 0.53  & 6.60  & 15.19  & 71.85 & 24.34 \\
    \bottomrule
    \end{tabular}%

}
  \label{tab:detail_b_q}%
\end{table}%

\paragraph{Typical Case.} Table \ref{tab:case} presents a typical case of {\ModelName-distill} in the third turn, comparing to Direct, and also the original response in the dataset. To better illustrate the effectiveness of the value learning, we present the reward and value scores for each response. In this case, {\ModelName} does not simply maximize the immediate reward, but maximizes the long-term return (\textit{i.e.}, the value), which is calculated from subsequent turns. Also, {\ModelName} in this case exhibits a perfect stage-turnover, guiding the conversation from the first stage (strategy Res.\&Par.), then the second stage (strategy Ref.), to the third stage (strategy Aff.\& Rea.). Comparing to Direct (stays in I) the original response (I to II to II), planning of {\ModelName} is more consistent with the theory proposed in \cite{liu2021ESconv}.

\section{Related Work}

There are some RL studies in goal-oriented conversations \cite{liDialogueActionTokens2024a,zhouArCHerTrainingLanguage2024a,liDQHGANHeterogeneousGraph2024}. For example, DAT \cite{liDialogueActionTokens2024a} defines dialogue action tags, and then generates responses by multi-turn planning. ArCHer \cite{zhouArCHerTrainingLanguage2024a} proposes a hierarchical RL algorithm to improve the efficiency and performance of LLMs. These works adapt the conversational LLM, and rely on ground truth annotations. In contrast, our {\ModelName} implements an explicit, lightweight and plug-and-play planner, which balances the foundation capability and the strategic thinking.

%learns to plan multi-turn explicit strategies offline from scoring feedback, and is decoupled from the dialogue model, which facilitates training and deployment.. 

% usually do not have a clear explicit strategy, and their evaluation often relies on ground truth values. 
% in multi-turn goal-oriented decision-making tasks
% represents conversations in a structured way by defining dialogue action tags and a multi-turn planner, and then generates responses based on these tags and pre-set planning strategies. 
%Typically, strategy learning occurs in goal-oriented conversations. Representative 
% , and  Q-star \cite{wangImprovingMultistepReasoning2024a} improves multi-step reasoning for LLMs
%  In addition, DQ-HGAN \cite{liDQHGANHeterogeneousGraph2024} realizes emotional support through the Graph Attention Network.

\section{Conclusion}
In this paper, based on Q-learning, we propose a method named \ModelName that optimizes long-term returns in emotional support conversation scenarios. Our implementation behaves as a plug-and-play strategic planner which steers the subsequent response generation. We propose two reward mechanisms, \ModelName-imit and \ModelName-distill, in which the former has higher automatic evaluation results, and the latter performs better in generalization and human preference alignment.

%We train to obtain the \ModelName-imit model, which has a higher strategy accuracy and a higher response consistency on the Esconv dataset. In addition, we proposes the \ModelName-distill method, which is trained from the reward signals of GPT-4's scoring and achieves more excellent human evaluation results, and can be generalized to a broader range of scenarios.

\clearpage
\newpage

\section{Limitation}
There are still some limitations of \ModelName. The results of the human evaluation may be biased, or deviate from the judgments of actual help-seekers due to the awareness of being engaged in scoring. Then, the test set may be small. Although it has little impact on the comparison between automated and human evaluations, sample sizes for some sub-categories may be insufficient when conducting a detailed analysis.

\bibliography{custom}

\newpage

\appendix

%\section{Further Implementation Details}

\section{Dataset details}
\label{sec:dataset}

\paragraph{Emotional Support Conversations.} Emotional Support Conversation (ESC) is a task aimed at alleviating users' negative emotions (e.g., anxiety, depression), where \textbf{supporters} assist \textbf{seekers} in managing emotions triggered by issues like work crises or interpersonal conflicts. Unlike emotion recognition tasks, ESC integrates psychological counseling mechanisms into the dialogue generation process, offering a deeper, context-sensitive solution for emotion regulation. The ESC dataset generally has the following attributes:

\begin{itemize}
    \item \textbf{Emotion}: Including emotion types and intensities, which help accurately capture the psychological state of the help-seeker.
    \item \textbf{Help-seeker profile}: A brief survey before each conversation that provides insight into the current situation of the help-seeker, revealing the challenges they are facing.
    \item \textbf{Situation}: A brief survey before each conversation that provides insight into the current situation of the help-seeker, revealing the challenges they are facing.
   \item \textbf{Strategy}: The response rule selected for the current turn based on the seeker’s emotional state. There are eight predefined rules in total.
   \item \textbf{Response}: The supporter’s response generated based on the history, inferred state, and selected rule.
\end{itemize}

In this paper, we mainly study two typical ESC datasets, ESConv and EmpatheticDialogues. ESConv has exactly the aforementioned architecture while EmpatheticDialogues lacks the Strategy. Table \ref{tab:statistics} summarizes the basic statistical information of ESConv and EmpatheticDialogues.

\begin{table}[ht!] 
  \centering
  \small
  \resizebox{0.49\textwidth}{!}{
    \begin{tabular}{llcc}
    \toprule
    %\multicolumn{3}{c}{\textbf{Statistics of ESConv}} \\ 
    \multicolumn{2}{c}{Category} & ESConv & EmpatheticDialogues   \\
    \midrule
    \multicolumn{2}{l}{\# Sessions} & 1.3K & 2.5K \\
    %\multicolumn{2}{l}{Average Session Length} & 543.6  \\
    \multicolumn{2}{l}{\# Uttr} & 38K & 11.0K\\
    \multicolumn{2}{l}{Average \# Uttr} & 28.9  & 4.3\\
    \multicolumn{2}{l}{Average Uttr Len} & 18.8 & 16.7\\
    \midrule
    \multirow{5}[0]{*}{Seeker} & \# Uttr & 20K & 5.7K \\
       & Avg \# Uttr & 15.4 & 2.2 \\
       & Avg Uttr Len & 16.8 & 20.8  \\
      %& \# Strategies & -& -\\
       & \# Emotions & 11 & 32 \\
    \midrule
    \multirow{5}[0]{*}{Supporter} & \# Uttr & 18K & 5.2K\\
       & Avg \# Uttr & 13.6 & 2.1  \\
       & Avg Uttr Len & 21.0 & 12.3  \\
       & \# Strategies & 8& -\\
       %& \# Emotions & -& 32 \\
    \bottomrule
    \end{tabular}%
  } % /Speaker1 /Speaker2
  \caption{
  Statistics of ESConv and EmpatheticDialogues. `Uttr' abbreviates Utterance. %For ESConv, we removed utterances from supporters at the beginning of dialogues because these utterances are usually uninformative greetings.
  }
  \label{tab:statistics}%
\end{table}

\begin{table}[t]
\centering
\small
\begin{tabular}{c|c}%{>{\centering\arraybackslash}m{0.25\textwidth}|p{0.7\textwidth}}
%\Xhline{1.2pt}
\toprule
%\multicolumn{1}{c|}{\textbf{Emotion Type}} & \multicolumn{1}{c}{\textbf{\# Occurrence}}  \\ 
Emotion Type & \# Occurrence \\
\toprule
anger & 111 \\ 
\midrule
anxiety & 354 \\ 
\midrule
depression & 334 \\ 
\midrule
disgust & 40 \\ 
\midrule
fear & 95 \\
\midrule
nervousness & 13 \\ 
\midrule
sadness & 308 \\ 
\midrule
shame & 42 \\ 
\bottomrule
%\Xhline{1.2pt}
\end{tabular}
\caption{Emotion statistics of ESConv.}
\label{tab:ESConv_emotions}
\end{table}

\paragraph{ESConv.} Motivated by the Helping Skills Theory \cite{Hill2009Helping}, \citet{liu2021ESconv} divides ESC into three sequential stages: \textit{Exploration}, \textit{Comforting}, and \textit{Action}, and proposes a dataset called ESConv. For each sample, the conversation is multi-turn, with the dialogue background and user emotion annotated. Upon each utterance of the supporter, 8 distinct support strategies are annotated. Table \ref{tab:ex_ESConv} exhibits an example of ESConv.

\begin{table*}[htbp!]

    \centering
    \small
    
    \begin{tabular}{c|l}
        \toprule
        \textit{\textcolor{green}{Topic}} & \makecell[l]{\textcolor{green}{I hate my job but I am scared to quit and seek a} \textcolor{green}{new career.}} \\
        \midrule
        \textit{Query} & \makecell[l]{\textit{\{history\}} \\
        \textit{seeker:} Seriously!\\ What I'm scare of now is how to secure another job.}  \\
\midrule
        \textit{\textcolor{red}{Emotion}} & {\textcolor{red}{Anxiety} (intensity: 5)} \\
        \midrule
       \textit{\textcolor{blue}{Strategy}} & \makecell[l]{\textcolor{blue}{Reflection of feelings}}   \\
\midrule
      \textit{Response} & \makecell[l]{\textit{supporter:} I can feel your pain just by chatting with you.}   \\
        \bottomrule
    \end{tabular}
    %\vspace{0.5cm}
    \caption{An example of \textit{ESconv}.} %  `Strategy' is the annotated ground truth.
    \label{tab:ex_ESConv}
\end{table*}

\begin{table*}[h!]
\centering
\small
\setlength{\tabcolsep}{8pt} % 调整列间距
\begin{tabular}{>{\centering\arraybackslash}m{0.15\textwidth}|m{0.05\textwidth}|m{0.65\textwidth}}
\toprule
\multicolumn{1}{c|}{\textbf{Strategies}} & \multicolumn{1}{c|}{\textbf{Abbr.}} &\multicolumn{1}{c}{\textbf{Definitions}}  \\ 
\hline
Question & \bf Que.& Inquiring about problem-related information to help the seeker clarify their issues, using open-ended questions for best results and closed questions for specific details. \\ 
\hline
Restatement or Paraphrasing &\bf Res.\& Par.& A simple, more concise rephrasing of the help-seeker’s statements that could help them see their situation more clearly. \\ 
\hline
Reflection of Feelings &\bf Ref.& Articulate and describe the help-seeker’s feelings. \\ 
\hline
Self-disclosure &\bf Self-Dis.& Divulge similar experiences that you have had or emotions that you share with the help-seeker to express your empathy. \\ 
\hline
Affirmation and Reassurance &\bf Aff.\& Rea.& Affirm the help seeker’s strengths, motivation, and capabilities and provide reassurance and encouragement. \\
\hline
Providing Suggestions &\bf Pro.& Provide suggestions about how to change, but be careful to not overstep and tell them what to do. \\ 
\hline
Information & \bf Inf.& Provide useful information to the help-seeker, for example with data, facts, opinions, resources, or by answering questions. \\ 
\hline
Others &\bf Others& Exchange pleasantries and use other support strategies that do not fall into the above categories. \\ 
\bottomrule
\end{tabular}
\caption{Strategy names, abbreviations and detailed definitions in ESConv.}
\label{tab:ESConv_strategies}
\end{table*}

Table \ref{tab:ESConv_strategies} provides definitions of support strategies in ESConv. Table \ref{tab:ESConv_emotions} lists the emotion types and their occurrences in the dataset. The emotion types include anger, anxiety, depression, disgust, fear, nervousness, sadness, and shame. 

\paragraph{EmpatheticDialogues.} EmpatheticDialogues \cite{rashkin-etal-2019-towards} is a dataset that consists of empathetic conversations. It aims to help in the development of empathetic language models by providing a large number of dialogues that express empathy. 

\section{Detailed Prompts}
\label{sec:detailed_prompt}

\paragraph{Instruction template.} To further strengthen the understanding capability of LLM on the strategy selection, we define the instruction as a multi-choice question (MCQ), forcing the LLM to choose one of the option numbers, instead of a plain question. Below is the content of the instruction template $\mathcal{I}$:

\begin{tabular}{p{0.9\linewidth}}

\hline
    You are a psychological consultant providing support to a seeker. The seeker's basic situation is as follows:\\
    Emotion: \{$e$\}\\
    Description: \{$desc$\}\\
    %Problem: \{problem\}\\
    %Situation: \{situation\}\\

    Below is the conversation history between the seeker and the supporter:\\
    \{$h$\}\\

    The seeker's current query is:\\
    \{$query$\}\\

    Based on the above context, please select the most appropriate response strategy from the following options:\\
    %A: Question\\ B: Restatement or Paraphrasing\\ C: Reflection of feelings\\ D: Self-disclosure\\ E: Affirmation and Reassurance\\ F: Providing Suggestions\\ G: Information\\ H: Others. \\
    strategy \#(1) \{$a_1$\}\\
    ...\\
    strategy \#(k) \{$a_k$\}\\
    Please provide your selection in the format of (1) through (k). Your selection is:\\
\hline
\end{tabular}

\paragraph{Generation prompt.} Below is the prompt used by the conversational foundation LLM for the response generation:

\begin{tabular}{p{0.9\linewidth}}

\hline
    You are a psychological consultant providing support to a seeker. The seeker's basic situation is as follows:\\
    Emotion: \{$e$\}\\
    Description: \{$desc$\}\\
    %Problem: \{problem\}\\
    %Situation: \{situation\}\\

    Below is the conversation history between the seeker and the supporter:\\
    \{$h$\}\\

    The seeker's current query is:\\
    \{$query$\}\\

    The current response strategy is:\\
    \{$a$\}\\

    Based on the current response strategy and other information, please act as a supporter and provide the best response. Keep replies brief without additional pronouns or extra elements.\\
\hline
\end{tabular}

\paragraph{Prompt of GPT-4 for reward generation.} Below is our prompt of GPT-4 to generate the rewards for {\ModelName}-distill:

\begin{table}[]
    \centering
    \begin{tabular}{p{0.95\linewidth}}

\hline
    You are a psychological consultant providing support to a seeker. The seeker's basic situation is as follows:\\
    Emotion: \{$e$\}\\
    Description: \{$desc$\}\\
    %Problem: \{problem\}\\
    %Situation: \{situation\}\\

    Below is the conversation history between the seeker and the supporter:\\
    \{$h$\}\\

    The seeker's current query is:\\
    \{$query$\}\\

    Please evaluate whether the response is appropriate:\\
    %A: Question\\ B: Restatement or Paraphrasing\\ C: Reflection of feelings\\ D: Self-disclosure\\ E: Affirmation and Reassurance\\ F: Providing Suggestions\\ G: Information\\ H: Others. \\
    \{$resp$\}\\
    Based on the information above, evaluate whether the response is suitable. Please remember to respond with a single integer number from 1 to 5, where 1 indicates "not suitable" and 5 indicates "very suitable". Please also provide a brief explanation of your decision.\\
\hline
\caption{Template of GPT-4 scoring.}
\label{tab:gpt_score}
\end{tabular}
\end{table}

\section{Principle of Human Scoring}
\label{sec:huam_score_principle}

We start with the criteria proposed by \citet{kang-etal-2024-large}. The human evaluation is aimed to align with the ultimate purpose of ESC, the seeker's \textit{satisfaction}. To achieve this, the supporter's behavior can be further classified into the following criteria:

\noindent \textit{Acceptance}: Does the seeker accept without discomfort;

\noindent \textit{Effectiveness}: Is it helpful in shifting negative emotions or attitudes towards a positive direction; 

\noindent \textit{Sensitivity}: Does it take into consideration the general state of the seeker. Furthermore, to clarify the capability of LLMs to align strategy and responses, we include Alignment.

To achieve a more elaborate assessment, we consider three more dimensions addressing the generation quality:

\noindent \textit{Fluency}: the level of fluency of response.

\noindent \textit{Emotion}: the emotional intensity of response which could affect the seeker's emotional state.

\noindent \textit{Interesting}: Whether the response can arouse the seeker's interest and curiosity, presenting unique ideas, vivid expressions or engaging elements that capture the seeker's attention and make the interaction more appealing.

We engage our interns as human evaluators to rate the models according to these multiple aspects, namely Fluency, Emotion, Interesting, and Satisfaction, with Satisfaction covering Acceptance, Effective, Sensitivity, and Satisfaction itself. \\
Throughout this evaluation process, we strictly comply with international regulations and ethical norms, ensuring that all practices conform to the necessary guidelines regarding participant involvement and data integrity.\\
To guarantee the accuracy and reliability of the evaluation results, a pre-evaluation training program is meticulously designed and implemented. During this training, the evaluation criteria are clearly and systematically expounded. Moreover, detailed explanations and scoring rules corresponding to each score are provided. \\
Evaluators are required to independently evaluate each sample in strict accordance with the pre-established criteria. By adhering to these principles, the evaluation process maintains objectivity, standardization, and consistency, thus enhancing the overall quality and credibility of the evaluation results. \\
The detailed manual scoring criteria are as follows:
\begin{itemize}
\item Fluency:

1: The sentence is highly incoherent, making it extremely difficult to understand and failing to convey a meaningful idea.

2: The sentence has significant incoherence issues, with only parts of it making sense and struggling to form a complete thought.

3: The sentence contains some incoherence and occasional errors, but can still convey the general meaning to a certain extent.

4: The sentence is mostly fluent with only minor errors or slight awkwardness in expression, and effectively communicates the intended meaning.

5: Perfect. The sentence is completely fluent, free of any errors in grammar, punctuation, or expression, and clearly conveys the idea.

\item Emotion:

1: The emotional expression is extremely inappropriate and chaotic, not in line with the content, and may convey wrong emotions.

2: The emotional expression has obvious flaws, either too weak or exaggerated, and is disjointed from the content.

3: The emotional expression is average. It can convey basic emotions but lacks depth and has minor issues.

4: The emotional expression is good. It can effectively convey the intended emotion with an appropriate intensity and is well integrated with the content.

5: The emotional expression is excellent. It is rich, nuanced, and perfectly matches the content, capable of evoking a strong and appropriate emotional response.

\item Acceptance:

1: The response inescapably triggers emotional resistance.

2: The response is highly likely to trigger emotional resistance.

3: The response has a possibility of emotional resistance occurring.

4: The response rarely provokes emotional resistance.

5: The response has no occurrence of emotional resistance.

\item Effectiveness:

1:  The response actually worsens the seeker's emotional distress.

2: The response carries the risk of increasing stress levels, and this outcome varies depending on the individual user.

3: The response fails to alter the seeker's current emotional intensity and keeps it at the same level.

4: The response shows promise in calming the emotional intensity; however, it is overly complicated or ambiguous for the user to fully comprehend and utilize effectively.

5: The response appears to be highly effective in soothing the seeker's emotions and offers valuable and practical emotional support. 

\item Sensitivity:

1: The response renders inaccurate evaluations regarding the seeker's state.

2: The response is characterized by rash judgments, as it lacks adequate assessment and in-depth exploration of the seeker's state.

3: The response is formulated with a one-sided judgment and a limited exploration of the seeker's state.

4: The response demonstrates an understanding that only covers a part of the seeker's state.

5: The response precisely grasps the seeker's state and is appropriately tailored according to the seeker's actual situation.

\begin{table}[t]
    \centering
    \resizebox{\columnwidth}{!}
    {
    \begin{tabular}{lcccccc}
        \toprule
        \textbf{Strategy}  & \textbf{Count} & \textbf{Ratio} & \textbf{Max} & \textbf{Min} & \textbf{Avg} & \textbf{Median} \\
        \hline
        Que. & 2574 & 17.6\% & 5 & 1 & 3.54 & 4 \\
        Res.\& Par. & 981 & 6.7\% & 5 & 1 & 3.48 & 3 \\
        Ref. & 1253 & 8.6\% & 5 & 2 & 3.65 & 4 \\
         Self-Dis. & 1410 & 9.6\% & 5 & 2 & 3.48 & 3 \\
         Aff.\& Rea. & 2444 & 16.7\% & 5 & 1 & 3.76 & 4 \\
         Pro. & 2367 & 16.2\% & 5 & 1 & 3.77 & 4 \\
         Inf. & 995 & 6.8\% & 5 & 2 & 3.75 & 4 \\
         Others & 2600 & 17.8\% & 5 & 1 & 3.77 & 4 \\
        \hline
        Total & 14624 & 100.0\% & 5 & 1 & 3.67 & 4 \\
        \bottomrule
    \end{tabular}
    }
 \caption{
 \label{tab：statistics_gpt}
  Statistics of GPT-4 score.}
\end{table}

\item Alignment:

1: The response is in total contradiction to the predicted strategy.

2: The response has a minor deviation from the predicted strategy.

3: There is some ambiguity between the response and the predicted strategy.

4: The response largely matches the predicted strategy, yet it contains some ambiguous elements.

5: The response effectively makes itself consistent with the predicted strategy.

\item Satisfaction:

1: The response is extremely disappointing. It doesn't answer the question at all and is of no help.

2: The response is poor. It only gives a partial answer and leaves many doubts unresolved.

3: The response is average. It meets the basic requirements but isn't particularly outstanding.

4: The response is good. It answers the question clearly and provides some useful details.

5: The response is excellent. It not only answers the question perfectly but also offers valuable additional insights.

\end{itemize}

%\section{More Results}
\section{Details of GPT-4 Scoring}

Table \ref{tab：statistics_gpt} presents GPT-4 score statistics across different response strategies. The overall average score is 3.67, with a median of 4. The most frequently used strategies are Others (17.8\%), Questioning (17.6\%), and Affirmation \& Reasoning (16.7\%), while Restating \& Paraphrasing (6.7\%) and Information Providing (6.8\%) appear less often. In terms of average score, Providing Opinions, Others, and Affirmation \& Reasoning score the highest (all-around 3.76–3.77), whereas Restating \& Paraphrasing and Self-Disclosure have the lowest average scores (3.48).

\end{document}